\documentclass{article}
\usepackage{arxiv_GPD}

\usepackage{times}
\usepackage{soul}
\usepackage{url}
\usepackage[hidelinks]{hyperref}
\usepackage[utf8]{inputenc}
\usepackage[small]{caption}
\usepackage{graphicx}
\usepackage{amsmath}
\usepackage{amsthm}
\usepackage{amssymb}
\usepackage{booktabs}
\usepackage{algorithm}
\usepackage{algorithmic}
\usepackage[switch]{lineno}
\usepackage{algorithmic}
\usepackage{appendix}
\usepackage{float} 
\usepackage[labelformat=simple]{subcaption}

\usepackage{multirow}
\graphicspath{{./images/}}

\title{Few-shot Face Image Translation via GAN Prior Distillation}

\author{
Ruoyu Zhao$^1$
\and
Mingrui Zhu$^2$\and
Xiaoyu Wang$^3$
\And
Nannan Wang$^{4}$\footnote{Co-corresponding authors}
\affiliations
$^1$State Key Laboratory of Integrated Services Networks, Xidian University, Xi'an, China\\
$^2$School of Computer Science and Technology, University of Science and Technology of China\\
\emails
royzhao@stu.xidian.edu.cn,
mrzhu@xidian.edu.cn,
fanghuaxue@gmail.com,
nnwang@xidian.edu.cn
}

\begin{document}
\maketitle
\begin{abstract}
Face image translation has made notable progress in recent years. However, when training on limited data, the performance of existing approaches significantly declines. 
Although some studies have attempted to tackle this problem, they either failed to achieve the few-shot setting (less than 10) or can only get suboptimal results. 
In this paper, we propose \textbf{G}AN \textbf{P}rior \textbf{D}istillation (GPD) to enable effective few-shot face image translation. GPD contains two models: a teacher network with GAN Prior and a student network that fulfills end-to-end translation. 
Specifically, we adapt the teacher network trained on large-scale data in the source domain to the target domain with only a few samples, where it can learn the target domain's knowledge. 
Then, we can achieve few-shot augmentation by generating source domain and target domain images simultaneously with the same latent codes. We propose an anchor-based knowledge distillation module that can fully use the difference between the training and the augmented data to distill the knowledge of the teacher network into the student network. 
The trained student network achieves excellent generalization performance with the absorption of additional knowledge. Qualitative and quantitative experiments demonstrate that our method achieves superior results than state-of-the-art approaches in a few-shot setting.
\end{abstract}

\section{Introduction}

Generative adversarial networks (GANs) \cite{goodfellow2014generative} have achieved promising results in image generation \cite{karras2019style,karras2020analyzing}, translation \cite{isola2017image,lin2021drafting}, editing \cite{perarnau2016invertible,roich2022pivotal}, and other tasks. However, the data used for training generally need thousands to tens of thousands of images, and the shortage of training data will seriously affect the performance of GAN-based models, resulting in poor quality of the generated images. Therefore, the difficulty of data acquisition and the high cost of making large datasets limit the application of GAN-based face image translation methods in many scenarios.

Several studies \cite{benaim2018one,liu2019few,wang2020semi} have attempted to reduce the data requirement of the image translation task. However, they failed to achieve the strict few-shot setting (less than 10 training images). JoJoGAN \cite{chong2022jojogan} and StyleGAN-NADA \cite{gal2022stylegan} have achieved translating images under few-shot, even one-shot settings, but their overall effect is still limited.

\begin{figure}[t]
\centering
\includegraphics[width=1.0\linewidth]{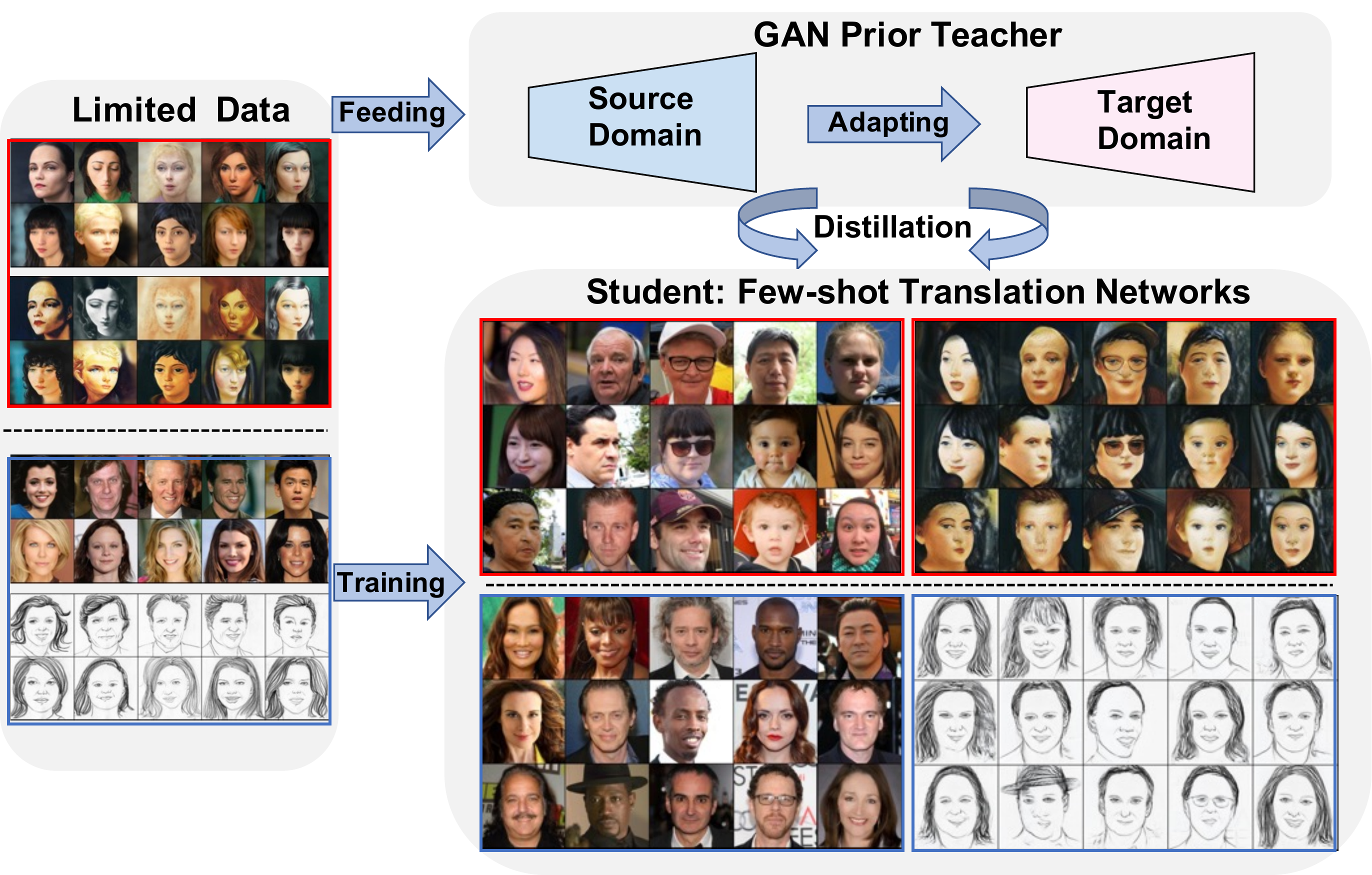}
\caption{The framework of our proposed method. We adapt the GAN prior teacher to the target domain to fulfill few-shot augmentation and distill the knowledge of the teacher network into the image translation (student) network in 10-shot setting.}
\label{fig:show}
\end{figure}

Recently, few-shot generative model adaption methods \cite{ojha2021few,xiao2022few,zhao2022closer} have successfully reduced the number of training images to less than 10 in the image generation task. They can get nontrivial results under the few-shot cross-domain generation setting. 
Their main idea is to adapt a pre-trained generation model \cite{karras2019style,karras2020analyzing} in the target domain with an auxiliary loss to preserve more source domain features. Thus the discriminator over-fitting is reduced, obtaining target images with better quality and well-preserved features.

Can the progress in the few-shot image generation task be extended to the few-shot image translation task? There are three challenges. (1) It is difficult to control the identity of the generated target image, and the identity can only be changed through sampling a latent code. 
(2) Although GAN inversion \cite{richardson2021encoding,tov2021designing} can be used to invert an image to control the identity, the identity and image quality will deteriorate in the process of inversion. On the other hand, it is hard to ensure that a high-quality target image can be obtained by inputting the inverted latent code into the target generator. 
Because it still has a gap between the inverted and the sampled latent vector. (3) Because the adaption process is unsupervised, which only uses the information of the style images, the results do not perform well on their test sets when applied to paired datasets. 
In addition, there are also some limitations of generation models themselves. (1) They usually contain a huge number of parameters, resulting in a large model size and long inference time. (2) The inversion speed is relatively slow. (3) All of the images used for inversion and generated by generation models need to be aligned. 
These three problems limit the application of the model in many practical scenarios.

We propose \textbf{G}AN \textbf{P}rior \textbf{D}istillation (GPD) to solve these problems and fulfill efficient and effective few-shot face image translation. The framework of GPD is shown in Figure \ref{fig:show}. GPD distills the knowledge of a teacher network, which has GAN prior in both the source and target domain, into an end-to-end face image translation (student) network. Although the training images are extremely limited (e.g., only 10 training image pairs are provided), the face image translation network can absorb additional knowledge about the source and target domains from the teacher network and obtain satisfactory translation results. GPD comprises two closely related modules: a few-shot generative augmentation module and an anchor-based knowledge distillation module. In the few-shot generative augmentation module, we first fine-tune the teacher network with only a few samples of the target domain to learn the target domain's knowledge so that it can generate target domain images. Then, we can achieve few-shot augmentation by generating source domain and target domain images simultaneously with the same latent codes. As for the anchor-based knowledge distillation module, we design a training strategy based on the anchor space. The limited training data is defined in this anchor space and the augmented data by the teacher network is defined outside it. Discriminators with different patch sizes (or receptive fields) are used when sampling data inside and outside the anchor space to adapt to the different levels of realism of the original data and augmented data. With this training strategy, the face image translation network can learn both the knowledge from the limited training data and the additional knowledge from the augmented data. Therefore, GPD achieves high generalization performance even in a few-shot setting.

The contributions of this paper are summarized as follows:
\begin{itemize}
\item We propose a GAN prior distillation framework for the challenging and practical few-shot face image translation task. GPD distills GAN prior into an end-to-end translation network, significantly improves the translation speed, and reduces the model's size.
\item We extend the few-shot generative model adaption and propose a few-shot generative augmentation module that can efficiently generate augmented image pairs.
\item We propose an anchor-based knowledge distillation module that leverages the difference between the training and the augmented data to distill the knowledge of the GAN prior teacher into the face image translation (student) network.
Extensive experiments demonstrate the superiority of our method over state-of-the-art methods in face image translation under few-shot settings.
\end{itemize}

\section{Related Work}
This section briefly reviews previous studies of few-shot image generation and image-to-image translation, which are most relative to our work.

\subsection{Few-shot Image Generation}
GANs excessively rely on massive training examples of high quality and diversity. For example, the satisfactory StyleGAN \cite{karras2019style,karras2020analyzing} is trained on a large dataset FFHQ containing 70k images. How to reduce the training data while ensuring the quality of results has always been a hot issue worth exploring.

An effective method is to use transfer learning to fine-tune the image generation model trained on a large dataset using a small amount of target data \cite{wang2018transferring,pinkney2020resolution,mo2020freeze}. It significantly reduces the training data and accelerates the convergence speed. However, it still needs thousands of training images.

To further reduce the training data, Elastic Weight Consolidation (EWC) is used in the fine-tuning process \cite{li2020few}. It evaluates the importance of each parameter and retains the relatively essential parameters during the tuning process to preserve more information of the pre-trained model. But the result is unsatisfactory in just a few training samples.

Recently, Cross-domain Correspondence (CDC) \cite{ojha2021few} has made a significant breakthrough in few-shot image generation. It uses an anchor-based strategy and a cross-domain distance consistency loss based on contrastive learning \cite{oord2018representation,chen2020simple} to adapt a pre-trained generation model, greatly reducing the discriminator overfitting under extremely few examples. 
RSSA \cite{xiao2022few} introduces a cross-domain spatial structural consistency loss to replace cross-domain distance consistency loss. It constrains the target domain to preserve more structure information of the source domain, thereby further reducing overfitting and achieving a better cross-domain generation effect than CDC. DCL \cite{zhao2022closer} and MixDL \cite{kong2022few} optimize the cross-domain distance consistency loss in CDC based on comparative learning. It also achieves better results than CDC.


\subsection{Few-shot Image-to-Image Translation}
Much research has provided solution schemes to train GANs with only a few data in recent years, but most of them are to solve classification problems. Because the results of classification tasks are relatively simple, having a higher false acceptance rate. The resulting form of the translation task is images, which is much more complex than classification probability. When training only under limited data, it is not easy to obtain results comparable to using large training sets.

Several methods have made efforts to reduce the training data of the image translation task. 
Compared with the traditional image-to-image translation tasks, OST \cite{benaim2018one} only uses one image in the source domain during training but still needs to use the image of the entire training set in the target domain.
FUNIT \cite{liu2019few} preliminary reduces the training data for unsupervised translation tasks. However, it cannot work well with extremely few data.
SEMIT \cite{wang2020semi} only requires limited labeled images during training but still needs a large amount of data to support training.

JoJoGAN \cite{chong2022jojogan} fine-tunes pre-trained StyleGAN \cite{karras2020analyzing} with style mixing and perceptual loss. Only one or a few target-style images are needed to be fine-tuned to change the result of the generated network into the target style. 
StyleGAN-NADA \cite{gal2022stylegan} proposed a CLIP-based loss that allows face image translation using text or a few image drivers. These methods support one-shot or even zero-shot, but the extent of style learning is limited.

\begin{figure}[t]
    \centering
    \includegraphics[width=1.0\linewidth]{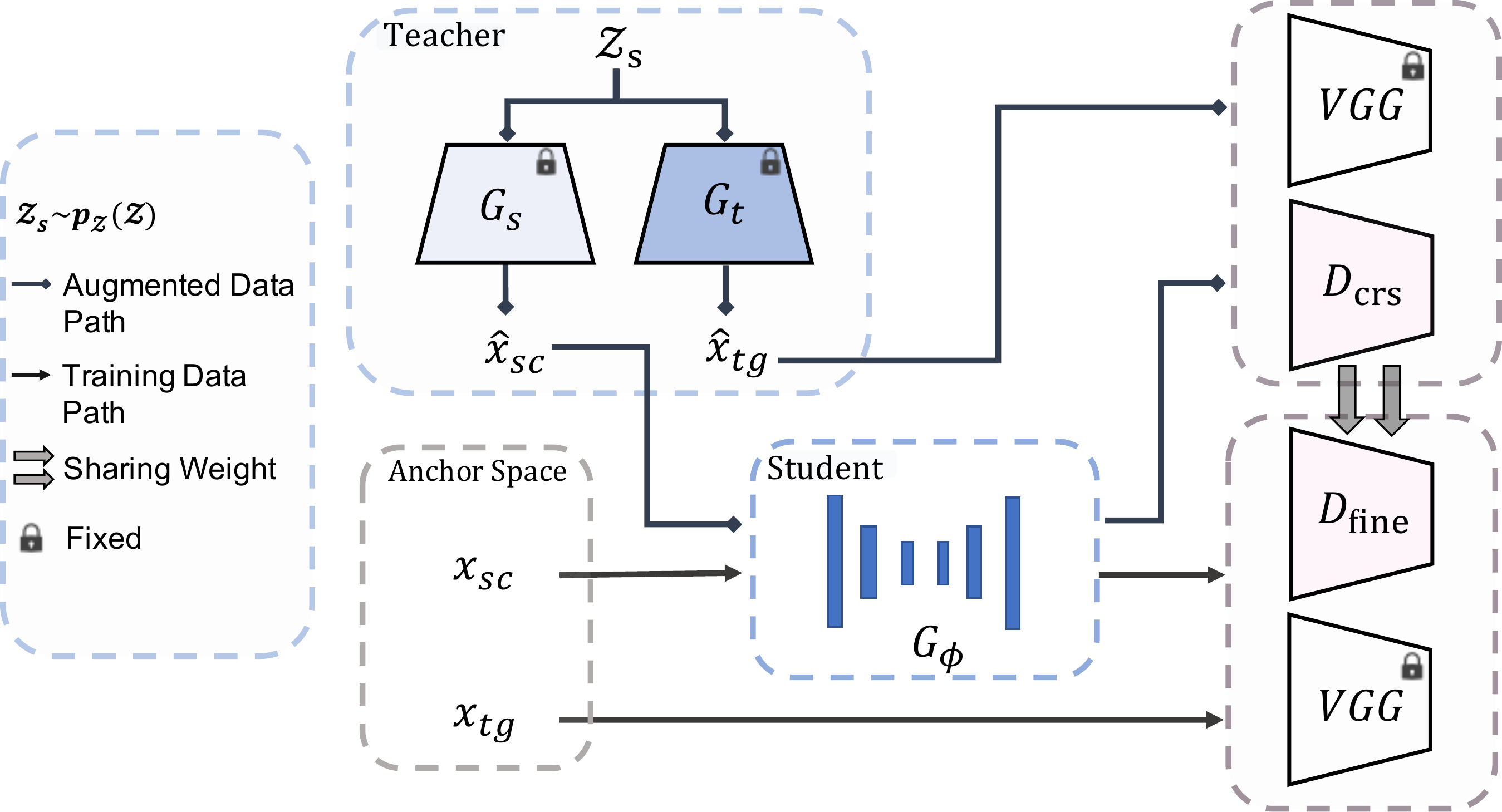}
    \caption{The total architecture of GPD. The network first adapts the teacher network with only a few samples,
            obtaining the target generator $G_t$ with the same structure as the source generator $G_s$.
            Then $G_s$ and $G_t$ are used to generate the augmented data pairs $\hat{x}_{\text{sc}}$, $\hat{x}_{\text{tg}}$ 
            to distill the student network together with the original training data pairs $x_{\text{sc}}$,  $x_{\text{tg}}$.
            Anchor space sampling and discriminators with different patch sizes are used for knowledge distillation. 
            $G_{\phi}$: student generator;  $D_{\text{crs}}$: student coarse-patch discriminator; $D_{\text{fine}}$: student fine-patch discriminator.}
    \label{fig:architecture}
\end{figure}

\section{Methodology}
Our goal is to distill the knowledge of the image generation (teacher) network into a face image translation (student) network to fulfill efficient and effective few-shot face image translation. 
Our proposed GPD comprises two highly correlated modules: a few-shot generative augmentation module and an anchor-based knowledge distillation module. In the former module, we use the existing few-shot generative model adaption method to adapt a teacher model trained on the large-scale dataset based on only a few data.
Thus, the teacher model can generate augmented image pairs for subsequent distillation.
In the latter module, we propose an anchor-based training strategy to distill the knowledge of the teacher network into the student network. It consists of two parts: anchor space sampling and discriminators with different patch sizes. 
The whole structure of GPD is shown in Figure \ref{fig:architecture}.

\subsection{Few-shot Generative Augmentation}
It is challenging to train a face image translation network using extremely few paired portraits, e.g., 10 pairs. Since the image generation network trained on a large dataset has achieved remarkable success in the generation task, we aim to extend it to train the few-shot image translation network. 
Specifically, we use the powerful image generation network as the teacher,
expanding the training data to get augmented data, then using two different data to train the student network. This can maximize the freedom of the number of images in the training set of the image translation network.

The selected teacher network should meet the following requirements. 
(1) It can generate images from different domains. Since the image generated by the teacher network needs to feed into the image translation network as the augmented image, the teacher network must be able to generate images from two domains. 
(2) It must be a few-shot image generation network. Because the training images are limited, the teacher network used for augmentation cannot be a network that requires a large number of training images, even though their generation effect is usually much better than that of few-shot generation networks. 
(3) The style of the generated image should be consistent with that of the training set as much as possible. 
We adapt a generation network pre-trained in the source domain to the target domain using existing few-shot image generation adaption methods based on limited training data. Then we extend it to augment paired data. Thus the target generator $G_t$ can generate target images while the source generator $G_s$ generates corresponding photos. 
When we distill the teacher's knowledge into the student network, the vector noise $z_s$ is sampled from the latent space to image generation networks $G_s$ and $G_t$ to generate paired images $\hat{x}_{\text{sc}}, \,\hat{x}_{\text{tg}} \in \mathcal{X}_{\text{aug}}$. This process can be expressed as below:

\begin{equation}
\begin{aligned}
\hat{x}_{\text{sc}} = {G}_{s}(z_s), \,z_s \sim p(z) \subset \mathcal{Z},
 \\
 \hat{x}_{\text{tg}} = {G}_{t}(z_s), \,z_s \sim p(z) \subset \mathcal{Z}.
\end{aligned}
\end{equation}
Following this way, the extremely limited training data can be augmented using the knowledge of the pre-trained generation network.

\subsection{Anchor-based Knowledge Distillation}
A naive method to distill the knowledge of the source and the target generation networks into the face image translation network ${G}_{\phi}$ is only using the augmented data pairs in training. The adversarial optimization objective is as below:
\begin{equation}
\begin{aligned}
\mathcal{L}_{\text{adv}} \left(G_{\phi}, D\right) &= \mathbb{E}_{\hat{x}_{\text{tg}} \sim \mathcal{X}_{\text{aug}}} \big[\left( D \left(\hat{x}_{\text {tg}} \right) \right)^2 \big] \\
&+ \mathbb{E}_{\hat{x}_{\text{sc}} \sim \mathcal{X}_{\text{aug}}} \big[ \left(1 - D \left({G}_{\phi} \left( \hat{x}_{\text {sc}} \right) \right) \right)^{2} \big].
\end{aligned}
\end{equation}%

In addition to adversarial loss, we use the perceptual loss \cite{johnson2016perceptual} to constrain the structure of the output images. We extract the feature maps from the i-th layer activations of a pre-trained VGGNet \cite{simonyan2014very}, which is denoted as $\Phi_{i}(\cdot)$.
We use $C_j$, $H_j$ and $W_j$ to indicate channel numbers, height and width of the feature maps, respectively. The perceptual optimization objective is as below:
\begin{equation}
\begin{aligned}
\mathcal{L}_{\text {per}} = \mathbb{E}_{\hat{x}_{\text{sc}, \text{tg}} \sim \mathcal{X}_{\text{aug}}} \big[\frac{1}{C_j H_j W_j}\left\|\Phi_j({G}_{\phi}(\hat{x}_{\text{sc}}))-\Phi_j(\hat{x}_{\text{tg}})\right\|_1 \big].
\end{aligned}
\end{equation}%

However, not only the data augmented by the teacher network is affected by overfitting, but the style of the augmented data still has a gap with actual training data. 
The result of the image translation network only trained on augmented data is unsatisfactory, especially compared to the ground truth on the test set. 
We want to maximize the paired image similarity and high-quality image style of the original training data. Inspired by \cite{ojha2021few}, we propose an anchor-based knowledge distillation strategy.

It consists of two parts: anchor space sampling and discriminators with different patch sizes. 
To maximize the use of the properties of the original training data, we define the original training data pairs ${x}_{\text{sc}}, \,{x}_{\text{tg}} \in \mathcal{X}$ within the anchor space. 
In contrast, we define the augmented data pairs $\hat{x}_{\text{sc}}, \,\hat{x}_{\text{tg}} \in \mathcal{X}_{\text{aug}}$ that is relatively insufficient in paired image similarity and style quality outside of the anchor space. 
When we sample data in the narrow anchor space, limited samples are repeatedly selected. So the network will pay more attention to the training data. 
When we sample data beyond the anchor space, the randomly sampled noise vector is fed into the teacher network to generate an augmented data pair for training. 
It is advantageous to sample different noise vectors at each iter instead of always selecting a fixed number of noise vectors.
First, it weakens the effect of a single augmented data pair on the training process, avoiding the inferior quality of individual augmented data affecting the training effect.
Then, it improves the generalization ability of the student network because it learns more knowledge from the teacher.
Apart from anchor space sampling, because of the gap in the realism of the training data and the generated data, we utilize discriminators with different patch sizes to constrain the two types of data with different realism. 
Specifically, we use the fine-patch discriminator $D_{\text {fine }}$ for the original training data in the anchor space to follow its better style quality. 
Meanwhile, we use the coarse-patch discriminator $D_{\text {crs }}$ for the augmented data outside the anchor space to follow its unsatisfactory quality.
\begin{figure}[tp]
    \flushleft 
	\subcaptionbox{Comparison of U-Net with and without augmentation.\label{plot1}}{
		\includegraphics[width=1.0\linewidth]{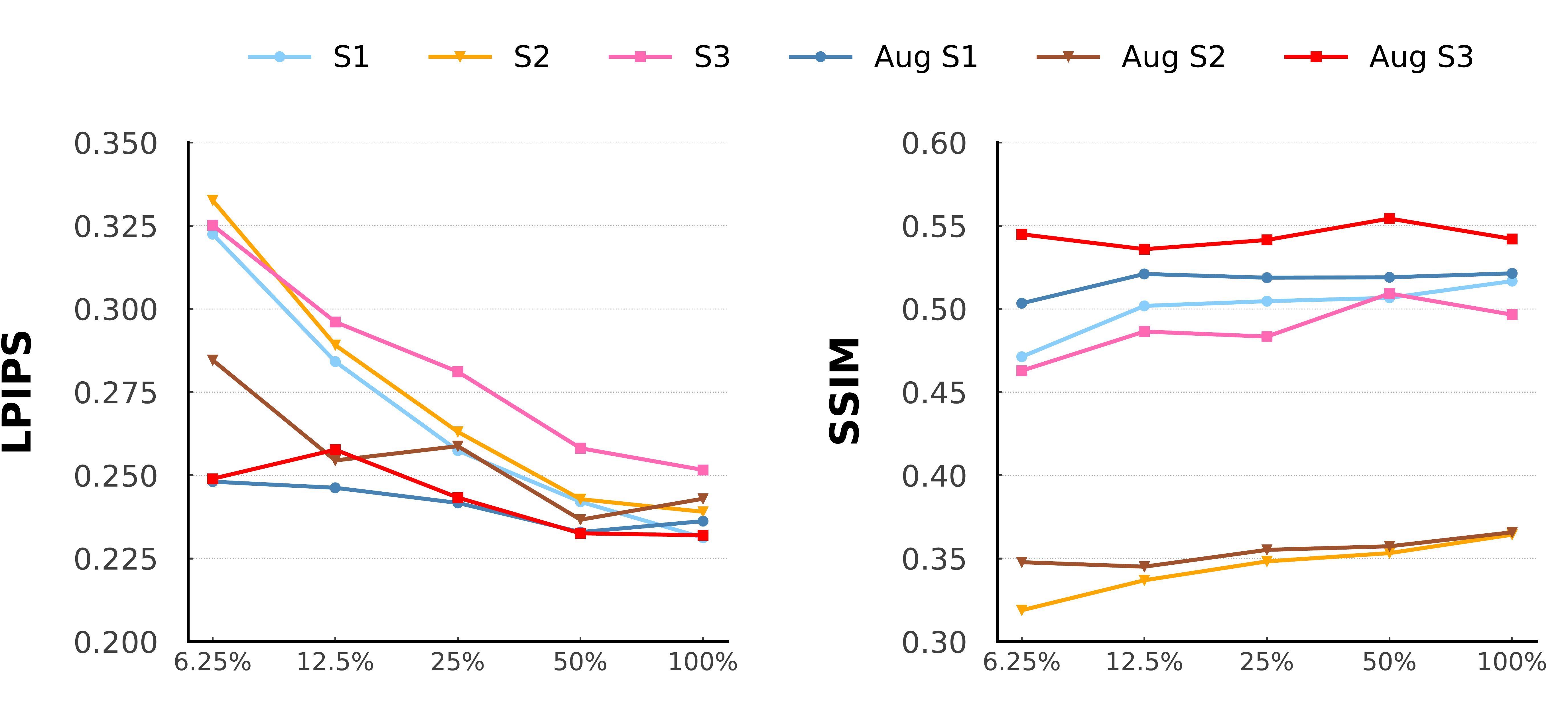}		
	}
	\hfill 
	\subcaptionbox{Comparison of Pix2Pix with and without augmentation.\label{plot2}}{
		\includegraphics[width=1.0\linewidth]{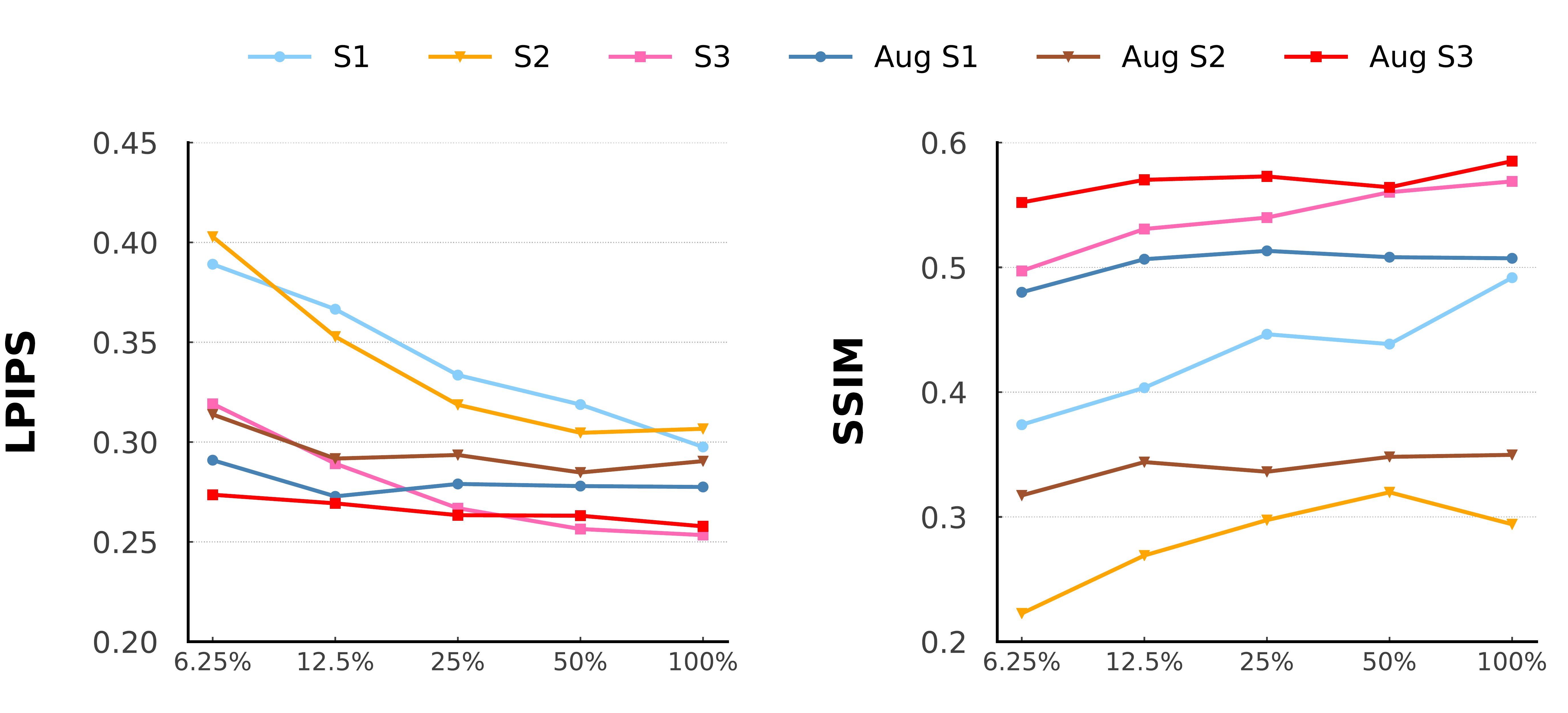}	
	}
	\caption{The LPIPS and SSIM comparison for U-net and Pix2Pix with and without augmentation under different data scales.}
	\label{fig:plot}
\end{figure}
In general, the adversarial optimization objective is as below:
\begin{equation}
\begin{aligned}
\mathcal{L}_{\text{adv}}^{\prime} \left({G}_{\phi}, \,{D_{\text{fine}}}, \,{D_{\text{crs}}}\right) &= \mathbb{E}_{{x}_{\text{sc},\,\text{tg}} \sim \mathcal{X}} \mathcal{L}_{\text{adv}} \left({G}_{\phi}, \,D_{\text{fine}}\right) \\
&+ \lambda_{1} \mathbb{E}_{\hat{x}_{\text{sc},\,\text{tg}} \sim \mathcal{X}_{\text{aug}}} \mathcal{L}_{\text{adv}} \left({G}_{\phi}, \,D_{\text {crs}}\right).
\end{aligned}
\end{equation}%
The perceptual optimization objective is as below:
\begin{equation}
\begin{aligned}
\mathcal{L}_{\text{per}}^{\prime} = \mathbb{E}_{{x}_{\text{sc},\, \text{tg}} \sim \mathcal{X}}\mathcal{L}_{\text{per}} + 
\lambda_{2}\mathbb{E}_{\hat{x}_{\text{sc},\, \text{tg}} \sim \mathcal{X}_{\text{aug}}}\mathcal{L}_{\text{per}}.
\end{aligned}
\end{equation}%
We define the final optimization objectives as below:
\begin{equation}
\begin{aligned}
\mathcal{L} =
\min _G \max _{D_{\text{fine}}, D_{\text{crs}}} 
\mathcal{L}_{\text{adv}}^{\prime} \left({G}_{\phi}, \,{D_{\text{fine}}}, \,{D_{\text{crs}}}\right) +
\mu\mathcal{L}_{\text{per}}^{\prime}.
\end{aligned}
\end{equation}%
For simplicity, $\lambda_1$ $\lambda_2$ are set to 1 and $\mu$ is set to 5 by default. We define the sampling frequency within and without anchor space as 1:2 by default for the few-shot setting.

\begin{table}[t]
    \centering
    \scalebox{0.95}{
    \begin{tabular}{l l c c c c c}
        \toprule
                     Style & FID$\downarrow$  &6.25$\%$  &12.5$\%$  &25$\%$ &50$\%$  &100$\%$ \\
        \midrule
        \multirow{2}*{S1} &BL     &0.0430 &0.0173 &0.0211 &\textbf{0.0023} &0.0067 \\
                          &Aug   &\textbf{0.0092} &\textbf{0.0043} &\textbf{0.0046} 
                                    &0.0039 &\textbf{0.0055} \\
        \midrule
        \multirow{2}*{S2} &BL     & 0.1200 & 0.0569 & \textbf{0.0349} & 0.0250 & 0.0370 \\
                          &Aug     & \textbf{0.0250}  & \textbf{0.0183} & 0.0561 & \textbf{0.0149} & \textbf{0.0336} \\
        \midrule
        \multirow{2}*{S3} &BL     & 0.0464 & 0.0136 & 0.0082 & \textbf{0.0063} & 0.0082 \\
                          &Aug     & \textbf{0.0101} & \textbf{0.0094} & \textbf{0.0064} &0.0068 & \textbf{0.0055} \\
        \bottomrule
    \end{tabular}}
    \caption{The FID comparison of the U-net with and without augmentation under different data scale. BL: baseline; Aug: using GPD for augmentation; S1, S2, S3: the three styles of FS2K.}
    \label{tab:augmentation}
\end{table}

\subsection{Overall Architecture}
GPD uses extremely limited data to distill GAN prior into an end-to-end translation network. 
In GPD, a few-shot generative augmentation module is proposed to extend the few-shot generative model adaption and an anchor-based knowledge distillation module is proposed to leverage the difference between the training data and augmented data.
In the former module, we use \cite{xiao2022few} to adapt the generation model \cite{karras2020analyzing} trained on the large-scale dataset. They work together as a teacher network to generate augmented image pairs efficiently.
For the student network, we use the lightweight U-Net based on \cite{zhu2021sketch} as the generator, which consists of the feature embedding encoder and the MSPADE-Decode. It allows faster translation speed and uses fewer computing resources. 
Based on the PatchGAN \cite{isola2017image} discriminator, we set the patch sizes of the coarse-patch discriminator and fine-patch discriminator to 4:1.

\section{Experiment}

We conduct experiments in two settings: few-shot data augmentation and few-shot face image translation. 
For the former setting, we use U-net \cite{zhu2021sketch} and Pix2Pix \cite{isola2017image} as the augmented baseline. For the latter setting, we compare the visual effect with three state-of-the-art networks: JoJoGAN \cite{chong2022jojogan}, StyleGAN-NADA (NADA) \cite{gal2022stylegan} and RSSA \cite{xiao2022few} with GAN inversion (Inv-RSSA). 
We also compare the evaluation metrics with three state-of-the-art face sketch synthesis networks: Sketch Transformer \cite{zhu2021sketch}, FSGAN \cite{fan2021deep}, and PANet \cite{nie2022unconstrained}.

For paired training data, we randomly sample CUFS \cite{wang2008face} (including AR, CUFS, and XM2VTS) and FS2K \cite{fan2021deep} (including three styles) to train our few-shot face image translation network. 
To enable our network to learn more styles, we use 10-shot art portraits \cite{yaniv2019face}, such as face paintings by Amedeo Modigliani, Moise Kisling, and Raphael. 
To get data pairs, we make a simple destylization. Specifically, we first invert the style image using an encoder-based method \cite{tov2021designing} to get the latent code $w^+$, and then we use style mixing to replace the upper layers with random noise to enhance realism. 
Finally, we use FFHQ \cite{karras2019style} to test the results. 

\begin{figure}[t]
    \centering
    \includegraphics[width=0.98\linewidth]{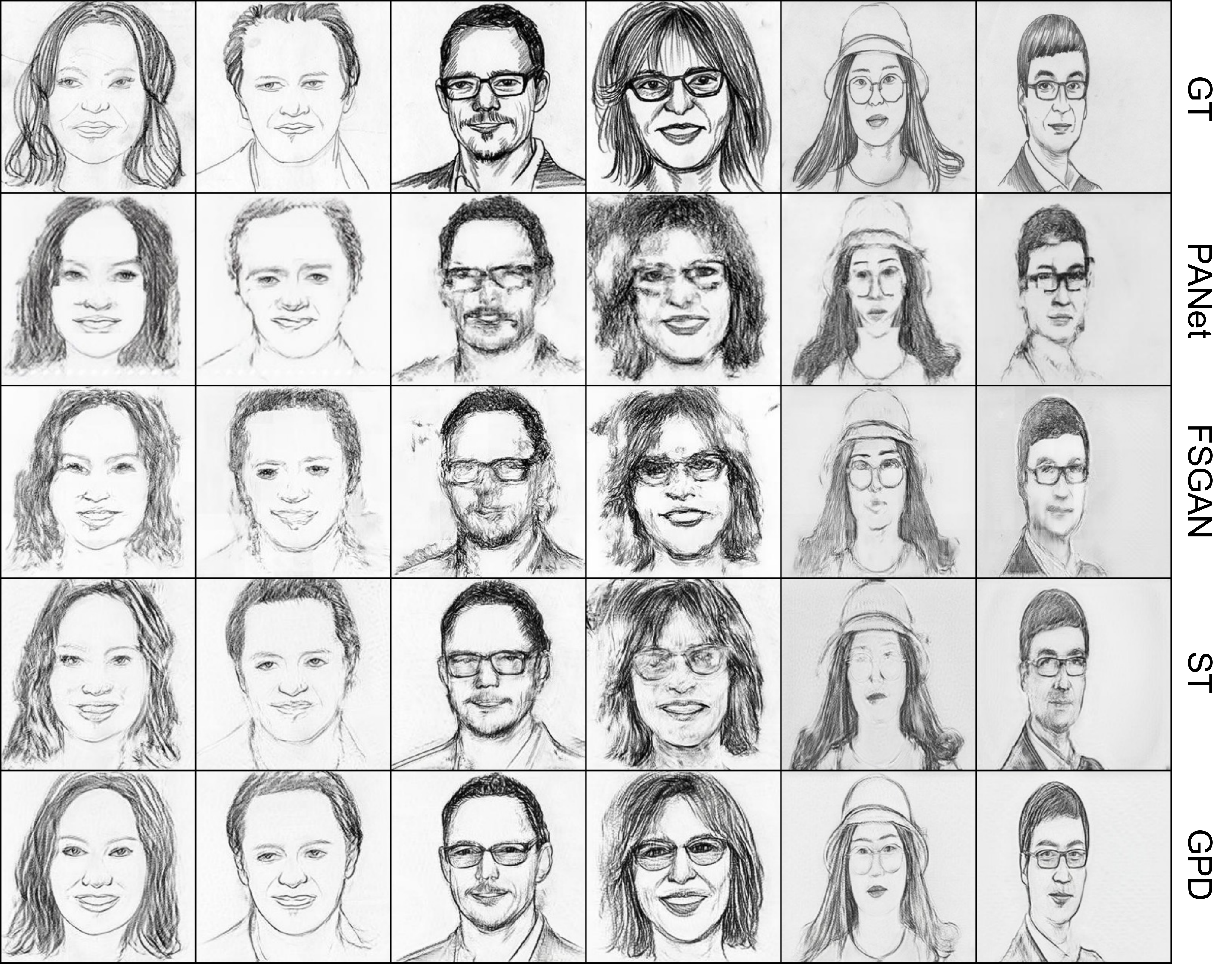}
    \caption{Results comparison with three sketch synthesis networks in 20-shot setting. We display two samples for each style.}
    \label{fig:compare_fss}
\end{figure}


\begin{figure*}[t]
    \centering
    \includegraphics[width=0.98\linewidth]{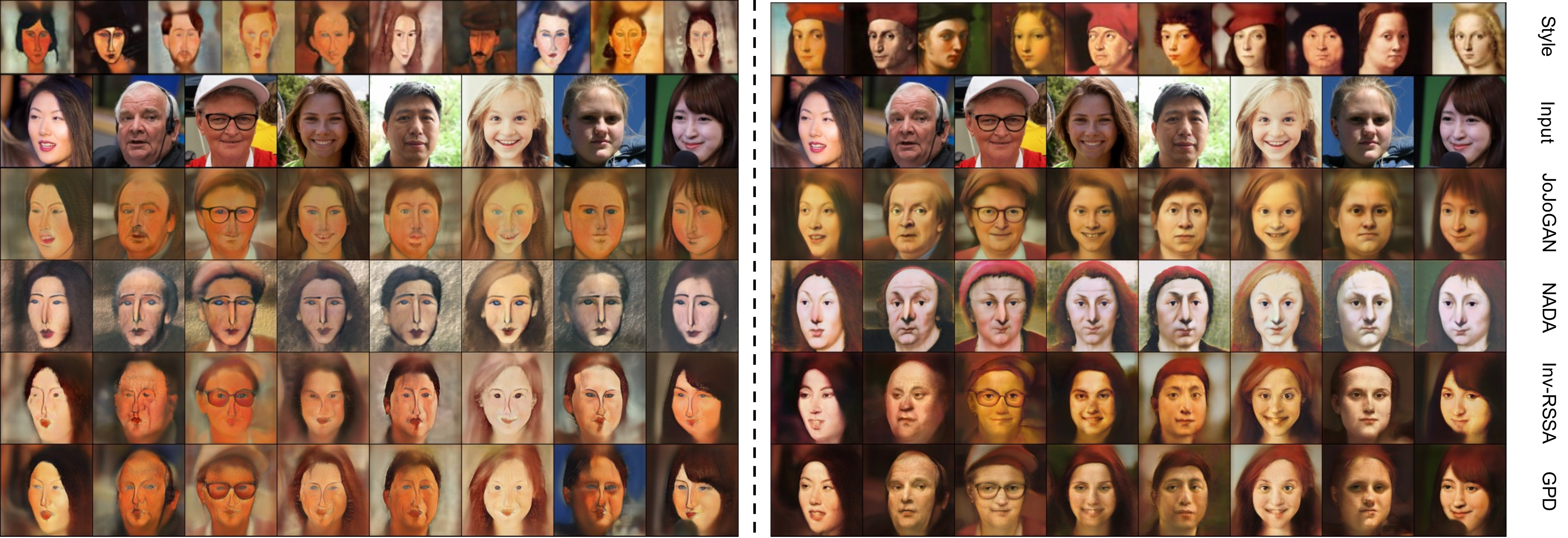}
    \caption{Comparison of 10-shot image translation result testing on FFHQ between different networks. The styles are the face paintings by Amedeo Modigliani (left) and the face paintings by Raphael (right). Zoom-in for details.}
    \label{fig:result1}
\end{figure*}

\begin{figure*}[t]
    \centering
    \includegraphics[width=0.98\linewidth]{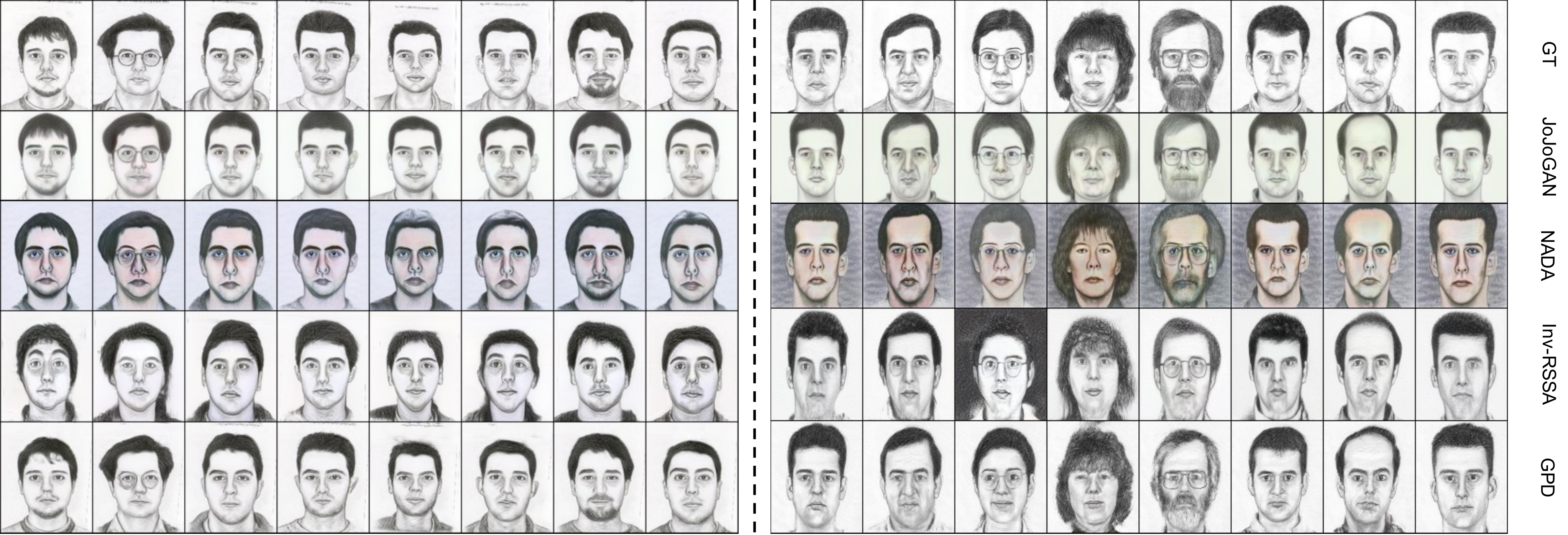}
    \caption{Comparison of 10-shot image translation result testing on their test sets between different networks. The styles are AR (left) and XM2VTS (right). Zoom-in for details.}
    \label{fig:result2}
\end{figure*}

\subsection{Few-shot Data Augmentation}

We extend the few-shot image generation network to augment limited training data. This method does not require a large number of training data. 
It still works even if the training data is less than 10. It is beyond the reach of the vast majority of other data augmentation methods. 

As shown in Figure \ref{fig:plot}, we analyze the performance of using GPD to augment the face image translation network. 
For each style of FS2K, we randomly sample the data pairs for training and set the data scale of each experiment to 100$\%$, 50$\%$, 25$\%$, 12.5$\%$, and 6.25$\%$ of the whole training dataset (reduced from about 360 to 20 data pairs). 
For the network settings, Both networks use our few-shot generative augmentation module and anchor space sampling in the anchor-based knowledge distillation module. 
U-Net additionally uses the setting of discriminators of different patches, while Pix2Pix does not use it but retains its original discriminator for simplicity. Both of them are trained with around 30k iterations. 
The difference in evaluation metrics is noticeable between using GPD and keeping the original setting.
With the continuous reduction of data size, using our method is more and more superior to the baseline in LPIPS \cite{zhang2018unreasonable} and SSIM \cite{wang2004image}. 
We can also see that with the data scale decreasing from more than 350 to 20, the evaluation metrics of using GPD for data augmentation changes flatly. It reflects the superiority of our method in demand for data scale and the effect of augmentation.

We also compare the generalization capability of the network with and without using our method for data augmentation. It tests on U-net as shown in Table \ref{tab:augmentation}. 
Under different data scales, lower FIDs \cite{heusel2017gans} are universally obtained by using our method for augmentation, which indicates that the model's generalization ability is improved by using our method for augmentation.

\begin{table}[tb]
    \centering
    \begin{tabular}{l c c c c c c}
        \toprule
          \multirow{2}*{Method} &\multicolumn{2}{c}{S1} &\multicolumn{2}{c}{S2} &\multicolumn{2}{c}{S3}\\\cmidrule(r){2-7}
        
               & Alex & VGG & Alex & VGG & Alex & VGG \\         
        \midrule
        PANet &0.357 &0.432 &0.376 &0.454 &0.288 &0.397  \\      
        \midrule
        FSGAN    &0.361 &0.406 &0.370 &0.430 &0.309 &0.400 \\    
        \midrule
        ST    &\underline{0.284} &\underline{0.360} &\underline{0.306} &\underline{0.383} &\underline{0.284} &\underline{0.384} \\           \midrule
        GPD  &\textbf{0.248} &\textbf{0.343} &\textbf{0.287} &\textbf{0.37}1 &\textbf{0.249} &\textbf{0.356} \\                      
        \bottomrule
    \end{tabular}
    \caption{The LPIPS comparison of different sketch synthesis methods in 20-shot setting. S1, S2, S3: the three styles of FS2K.}
    \label{tab:fss}
\end{table}

\begin{table}[tb]
    \centering
    \begin{tabular}{l c c c c}
        \toprule
        Method  &JoJoGAN &NADA  &Inv-RSSA  &GPD \\  
        \midrule
       Preference ($\%$)   &\underline{36.67}  &7.00 &6.87  &\textbf{49.47} \\                      
        \bottomrule
    \end{tabular}
    \caption{User preference result of four face image translation methods. We count the average preference of users for the results of five styles.}
    \label{tab:rs}
\end{table}

\subsection{Few-shot Face Image Translation}
\begin{figure}[tb]
    \flushleft
	\subcaptionbox{Ablation study of visual quality.\label{ablation1}}{
	\includegraphics[width=0.95\linewidth]{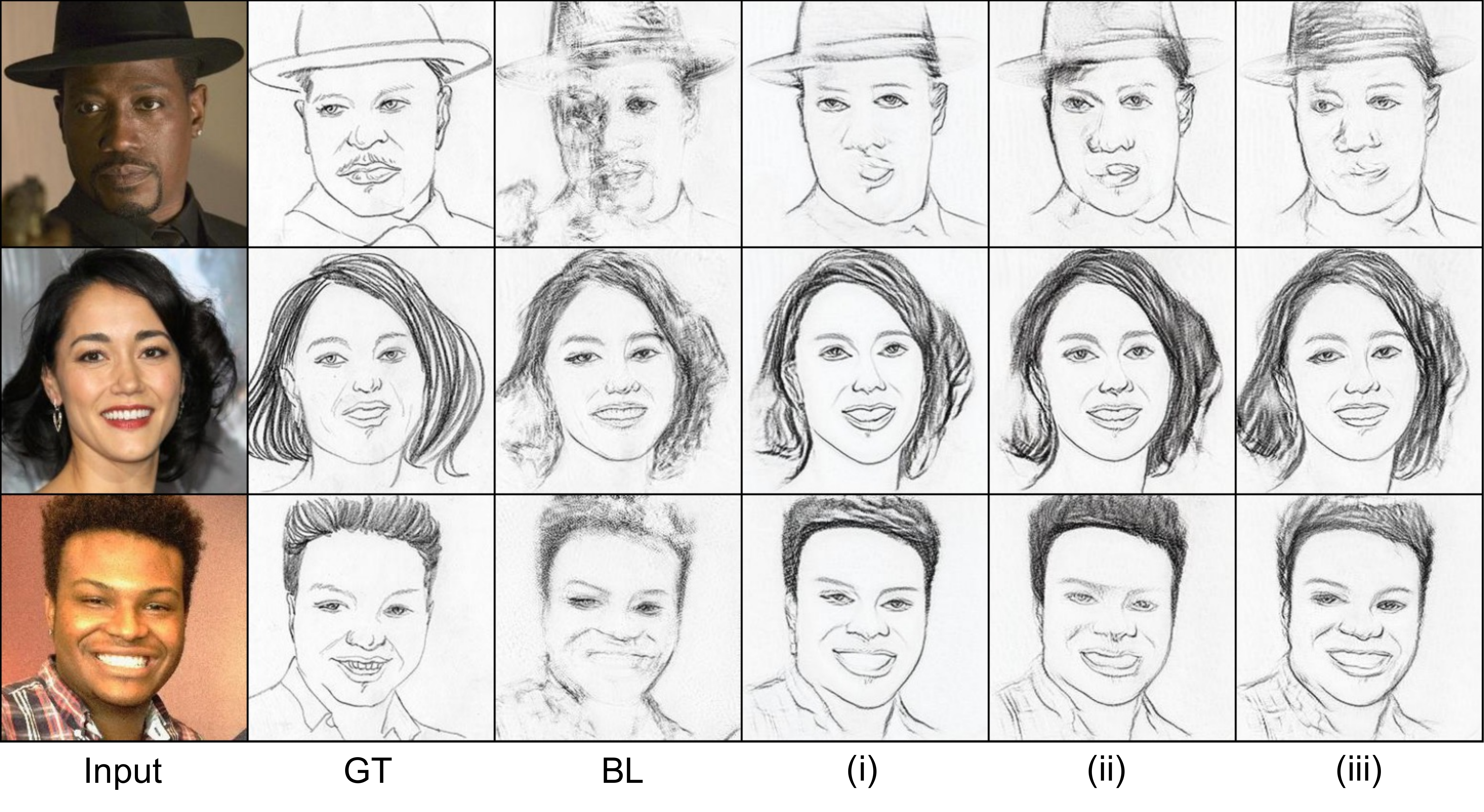}	
	}
	\hfill 
    \vspace{1em}
	\subcaptionbox{Ablation study of the overfitting extent.\label{ablation2}}{
	\includegraphics[width=0.95\linewidth]{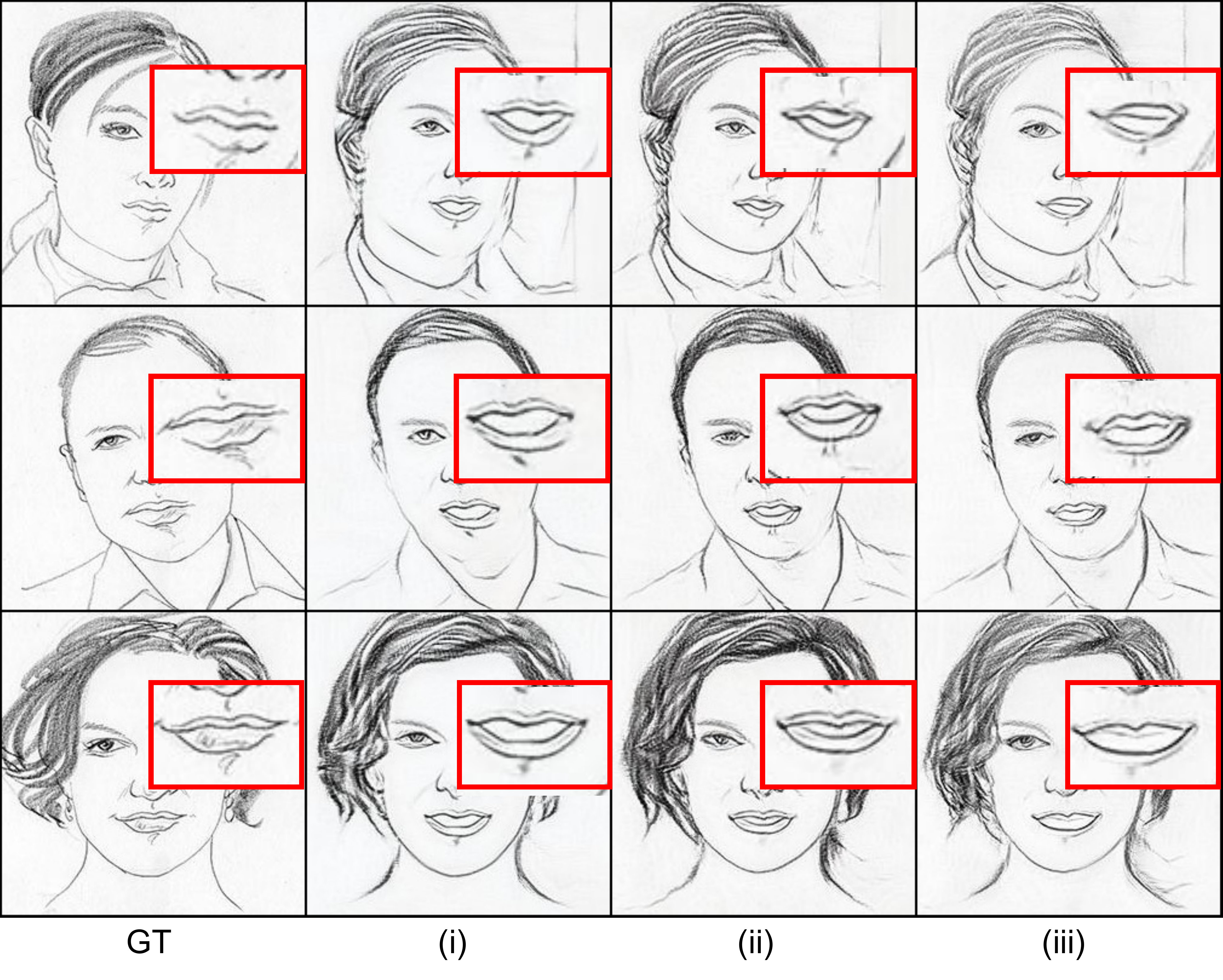}	
	}
	\caption{Ablation study of GPD in 20-shot setting. GT: ground truth; BL: baseline; (\romannumeral1): only with the augmented data; (\romannumeral2): with the anchor space sampling; (\romannumeral3): with the anchor space sampling and discriminators with different patch sizes (in full settings).}
	\label{fig:ablation}
\end{figure}

Based on the qualitative and quantitative comparison, we compare the results of our method with state-of-the-art methods. We train our network around 30k iterations. Without special instructions, the comparison methods are all based on their default settings. 

\textbf{Qualitative comparison.}
We compare our results with several methods of few-shot image translation, such as JoJoGAN and NADA. For NADA, we apply the image-driven approach to enable image-to-image translation.
As shown in Figure \ref{fig:result1}, JoJoGAN and NADA can only present suboptimal solutions when learning complex styles, which is worse than our method. NADA is more vulnerable to the influence of overfitting in few-shot tasks, which shows apparent local information overfitting in some regions. 
We also compare the performance with the few-shot image generation method RSSA \cite{xiao2022few} using GAN inversion for image translation. For translation accuracy, the best optimization target is the latent vector $z\subset \mathcal{Z}^+$, because the source and the target generator share the latent space $\mathcal{Z}$. 
Specifically, we invert the input image into a latent vector $z$ using the inversion method based on optimization \cite{abdal2019image2stylegan}. We add regularization  \cite{yang2022pastiche} to the loss to make the optimized $z$ closer to $\mathcal{Z}$ space. Then the latent vector $z$ is used as the input of the target generator to obtain the corresponding target image. 
Because there is still a gap between the $z$ optimized by inversion and the $z_0$ shared by the source domain and target domain, the generated effect cannot be guaranteed when $z$ is input into the target generator. This is particularly evident in the test set of the dataset itself, as shown in Figure \ref{fig:result2}. 
In addition, in the few-shot image generation method, the style image generated in the target generator is affected by overfitting, which is usually solved by setting additional constraints. GPD has better structure preservation than those adaption-based few-shot image generation methods because an image is input to the network as the prior.


\textbf{Quantitative comparison.}
Facial-sketch synthesis \cite{zhu2017deep,zhang2019neural,zhang2019deep} is a familiar face image translation task. 
We compare the LPIPS with three state-of-the-art facial-sketch synthesis networks, i.e., Sketch-Tranformer (ST), FSGAN, and PANet. 
The datasets we used are the three styles of FS2K in 20-shot setting. All of the methods are trained around 30k iterations.
GPD is superior to other facial-sketch image translation methods in few-shot settings. As shown in Table \ref{tab:fss}, we achieve optimal results in terms of the LPIPS. We also display the generation result of four networks, as shown in Figure \ref{fig:compare_fss}. GPD preserves a complete structure in few-shot settings.

We study the average preferences of 100 respondents for the result of four methods under five different datasets (face paintings by Amedeo Modigliani, Moise Kisling and Raphae, AR, and XM2VTS. There are 15 questions in total, and each data set corresponds to three questions), as shown in Table \ref{tab:rs}. The results show that our method is the optimum in visual effects.

\begin{table}[t]
    \centering
    \begin{tabular}{l l c c c c c}   
        \toprule
        \multirow{2}*{Style}  & \multirow{2}*{Setting} &\multicolumn{2}{c}{LPIPS} &        &         \\\cmidrule(r){3-4}
            & &Alex$\downarrow$ &VGG$\downarrow$ &SSIM$\uparrow$ &FSIM$\uparrow$ \\
        \midrule
        \multirow{3}*{S1} & BL   & 0.3225 & 0.3968 & 0.4713 & 0.6268  \\
                          & Aug   & \underline{0.2632} & \underline{0.3502} & \textbf{0.5132} & \underline{0.6323}  \\
                          & Aug\,+\,Anchor    & \textbf{0.2481} & \textbf{0.3431} & \underline{0.5034} & \textbf{0.6408}  \\
            
        \midrule
        \multirow{3}*{S2} & BL  & 0.3326 & 0.4133 & 0.3189 & \textbf{0.5923}  \\
                          & Aug   & \underline{0.2890} & \underline{0.3862} & \underline{0.3391} & 0.5704  \\
                          & Aug\,+\,Anchor    & \textbf{0.2846} & \textbf{0.3713} & \textbf{0.3478} & \underline{0.5897}  \\
                        
        \midrule
        \multirow{3}*{S3} & BL   & 0.3252 & 0.4161 & 0.4629 & 0.6115  \\
                          & Aug    & \underline{0.2634} & \underline{0.3765} & \underline{0.5379} & \underline{0.6310}  \\
                          & Aug\,+\,Anchor      &\textbf{0.2490} &\textbf{0.3562} &\textbf{0.5449} &\textbf{0.6455}  \\
                   
        \bottomrule
        \end{tabular}
    \caption{Ablation study under three styles of datasets in 20-shot setting. S1, S2, S3: the three styles of FS2K; BL: baseline; Aug: with few-shot generative augmentation module; Anchor: with anchor-based knowledge distillation module.}
    \label{tab:ablation}
\end{table}

\subsection{Ablation Analysing}
We respectively explore the impact of the two modules designed on the experimental results. 
Based on three styles of face sketch data pairs in FS2K, we randomly sample 20 pairs of data of each style, obtaining three styles of few-shot train sets for experiments. 
Then we use the test sets of FS2K to test the results. 

\textbf{Effect of the few-shot generative augmentation module.}
Using the teacher network for data augmentation can distill the knowledge of the large-scale generation network into a small-scale image translation network. 
When only augmented data is used for training, the student network can learn abundant knowledge of the teacher network. 
Therefore, it has much better structural integrity on the test sets than only using the original training set. Figure \ref{fig:ablation}\subref{ablation1} shows that BL is the baseline trained on 20-shot training data. 
Setting (\romannumeral1) is the setting where we only use the augmented data to train the network. 
It is better than BL in visual effect because its structural information is complete.
However, the student network excessively relies on augmented data under this setting. It is affected by the overfitting of the teacher network, as shown in Figure \ref{fig:ablation}\subref{ablation2}. 
In terms of evaluation metrics, data augmentation has dramatically improved the evaluation metrics of the student network, as presented in Table \ref{tab:ablation}.

\textbf{Effect of the anchor-based knowledge distillation module.}
The anchor-based knowledge distillation module can mitigate the influence of overfitting caused by the generated network.
As seen in Figure \ref{fig:ablation}\subref{ablation2}, when we only use the augmented data, the overfitting problem is severe (such as the mouth shape of most sketch images is almost the same). 
When we use the method of anchor space sampling, the extent of overfitting is further alleviated. When we apply discriminators with different patch sizes, the visual effects of the results are more diversified. 
Moreover, using the anchor space sampling and discriminators with different patch sizes progressively enhances the visual quality, as seen in Figure \ref{fig:ablation}\subref{ablation1}. 
In terms of evaluation metrics, using the anchor-based knowledge distillation module further improves the evaluation metrics, as shown in Table \ref{tab:ablation}.

\section{Conclusion}
We aim to use the powerful pre-trained GAN to generate data to help the training tasks with insufficient data. Therefore we propose GAN Prior Distillation (GPD). 
It can be applied to few-shot data augmentation and few-shot face image translation tasks. Our research is based on two essential modules: the few-shot generative augmentation and the anchor-based knowledge distillation module. 
Plenty of experiments demonstrate the advantages of GPD compared with the state-of-the-art methods, which promoted the progress of few-shot image translation and few-shot data augmentation.

GPD only works on face data because it extends the few-shot image generation adaption for data augmentation. Most few-shot image generation adaption is currently established on StyleGAN trained on the FFHQ face dataset. 
But our method can also be applied to other data with the new image generation model proposed.

\appendix




\bibliographystyle{named}
\bibliography{ijcai23}

\clearpage
\begin{appendices}
\section{Code}
We provide the code of our proposed GPD model. Please refer to the "README.md" file for detailed usage of the code.


\section{The Results of Few-shot Face Image Augmentation and Translation}
We provide the high-resolution results of few-shot face image augmentation and translation. 

\begin{figure*}[t]
        \centering
        \includegraphics[width=1\linewidth]{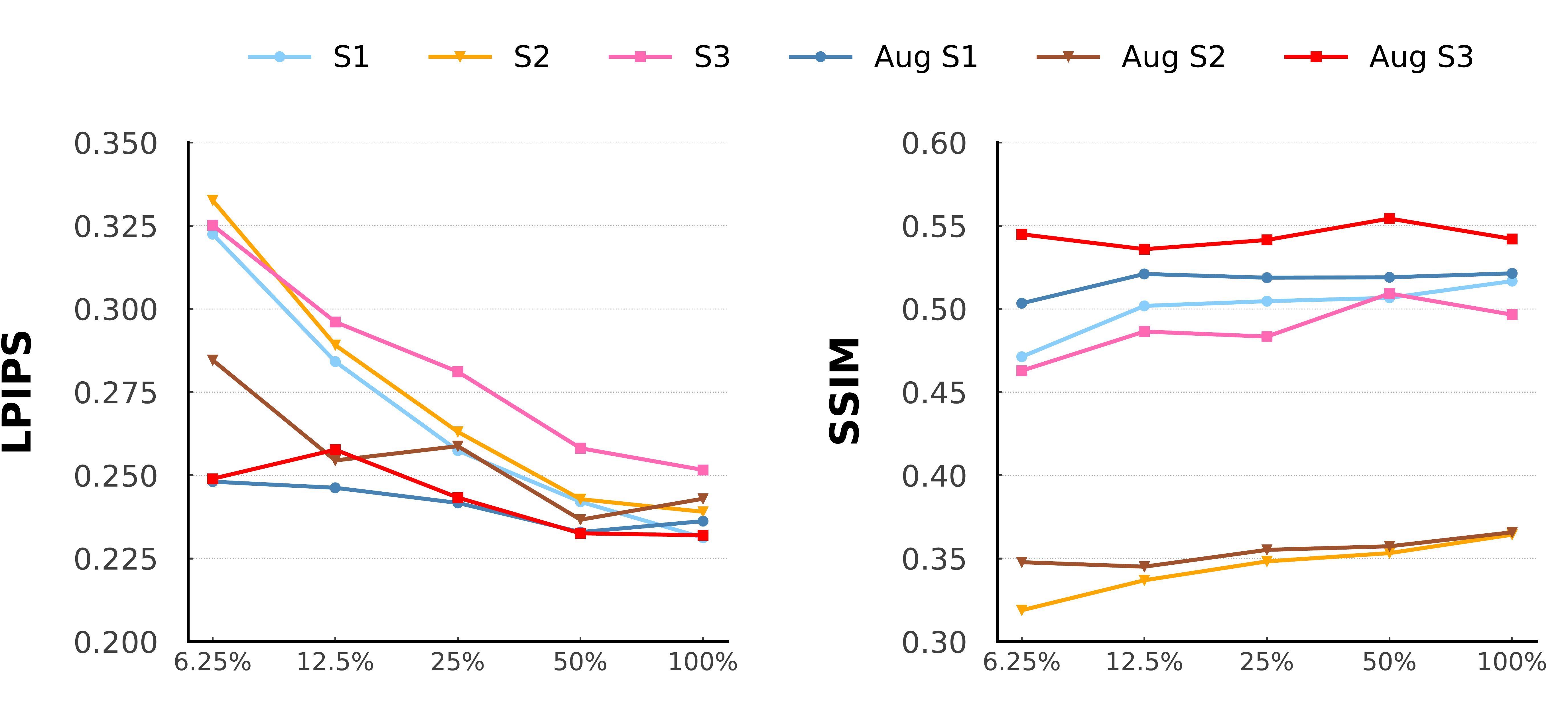}
        \caption{ The LPIPS and SSIM comparison for U-net
        with and without augmentation under different data scales.}
        \label{fig:ap_plot1}
        \end{figure*}

\begin{figure*}[t]
        \centering
        \includegraphics[width=1\linewidth]{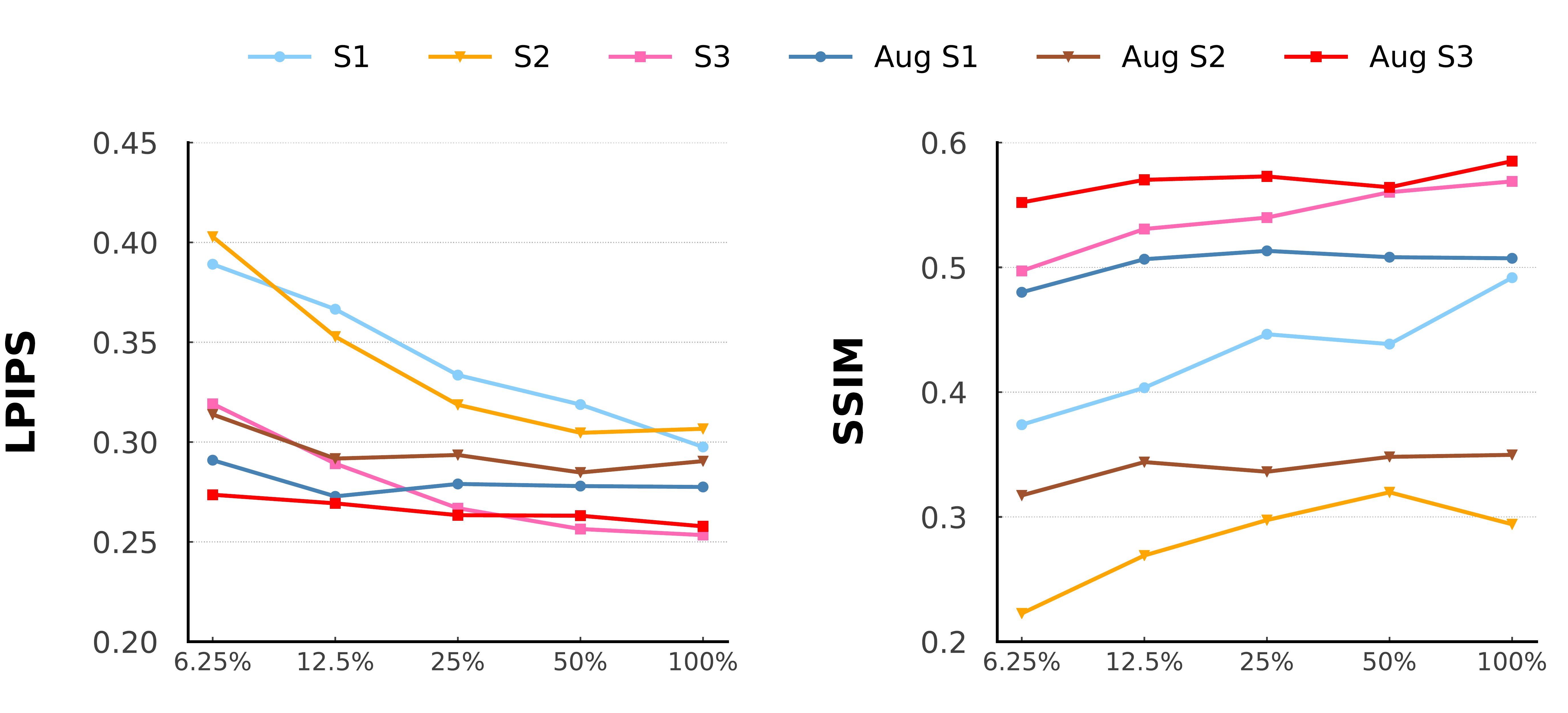}
        \caption{ The LPIPS and SSIM comparison for Pix2Pix
        with and without augmentation under different data scales.}
        \label{fig:ap_plot2}
        \end{figure*}

Table \ref{tab:ap_plot1}-\ref{tab:ap_plot2} are the results of the few-shot face image augmentation, showing the differences between LPIPS and SSIM results of U-Net and Pix2Pix networks with and without using our GPD for augmentation.
According to the tables, better LPIPS and SSIM are universally acquired with our augmentation method.
Figure \ref{fig:ap_plot1}-\ref{fig:ap_plot2} are the corresponding line charts, which more intuitively show the advantages of using our method to augment data.

\begin{figure*}[t]
    \centering
    \includegraphics[width=1.0\linewidth]{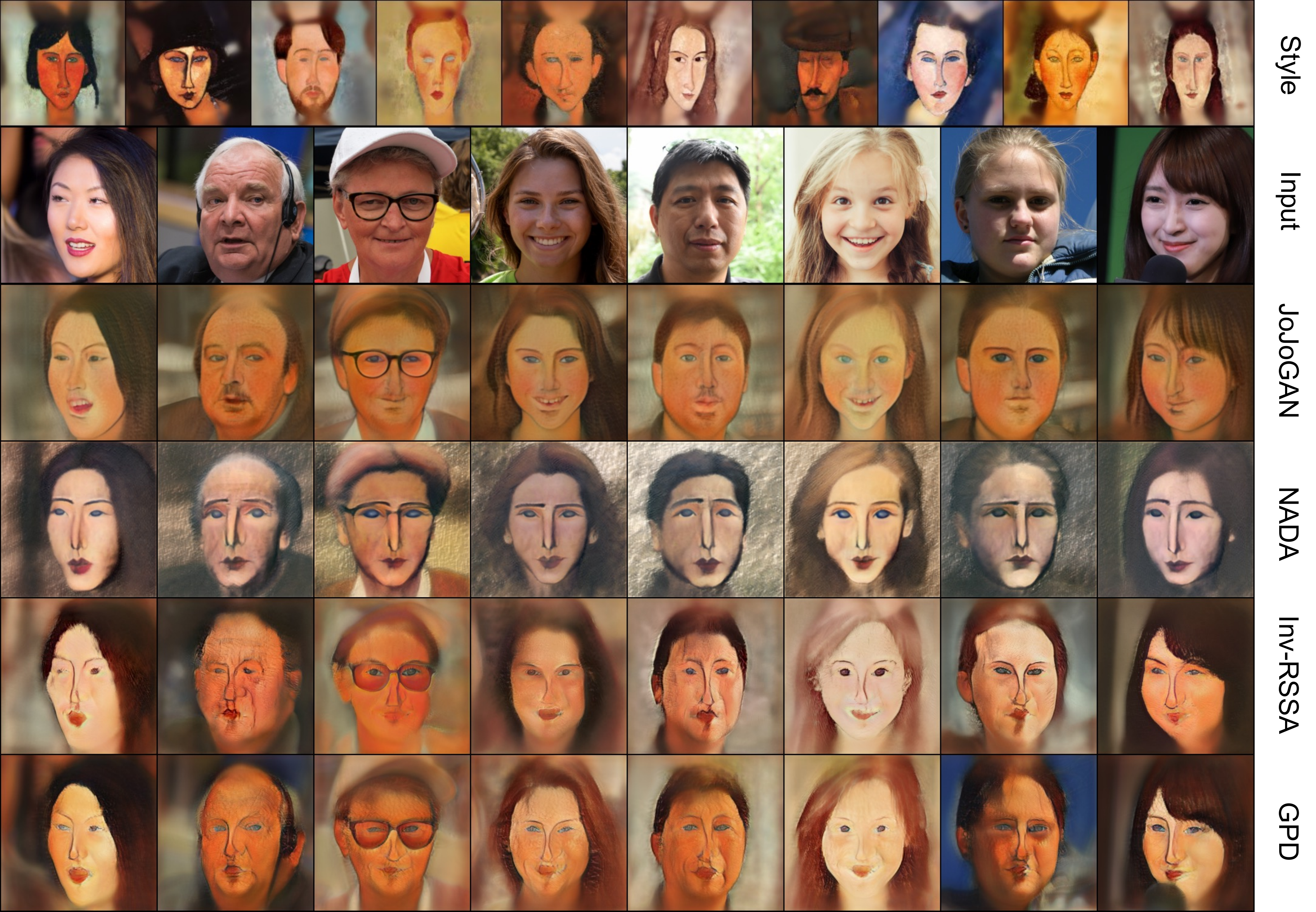}
    \caption{Results comparison on face paintings by Amedeo Modigliani in 10-shot setting. We test the results on FFHQ, which is the source domain of the other three models. GPD presents a competitive style compared to other methods and the most consistent content with the source image (Input). It also shows the generalization performance of GPD.}
    \label{fig:ap_a}
    \end{figure*}

\begin{figure*}[t]
    \centering
    \includegraphics[width=1.0\linewidth]{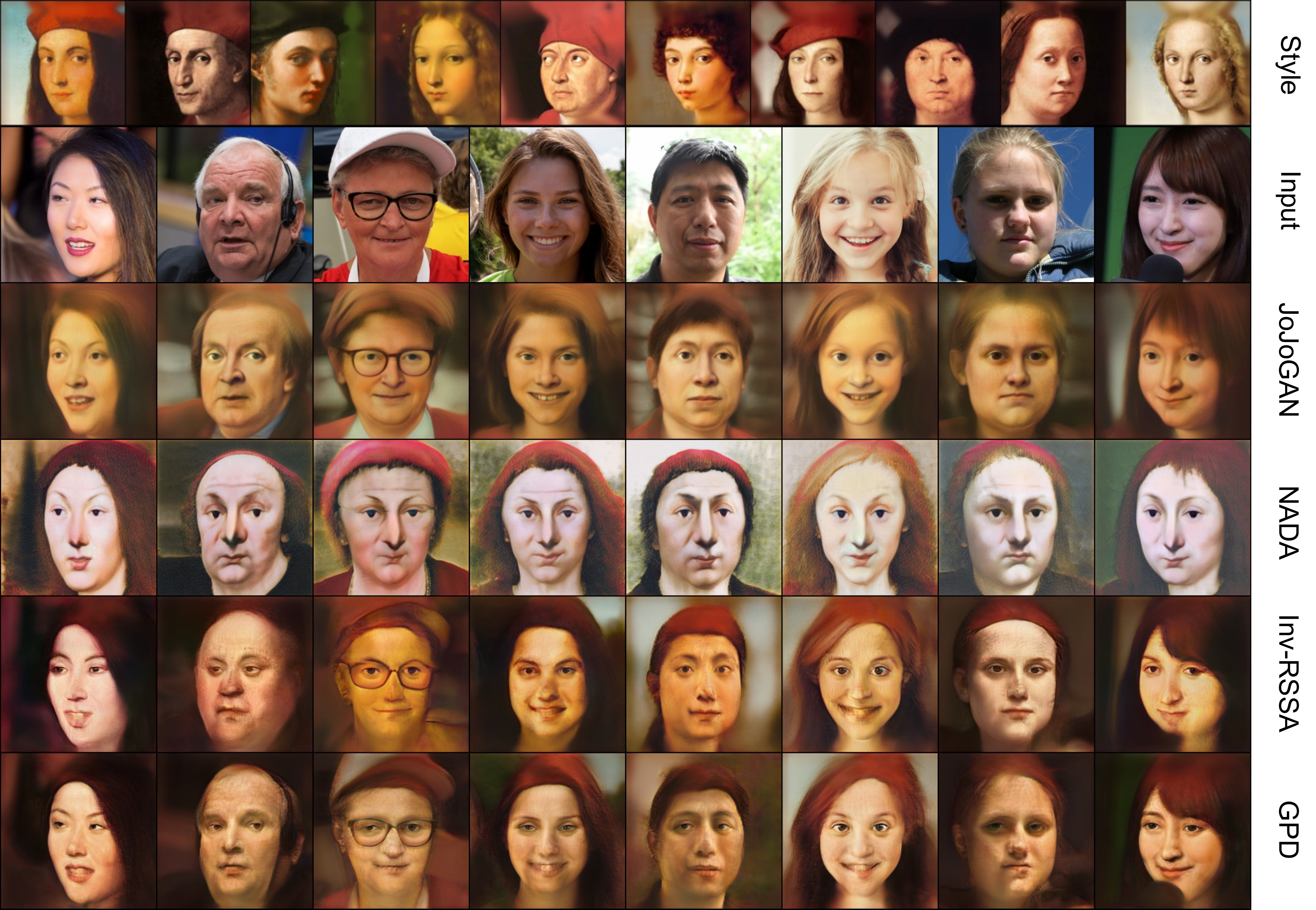}
    \caption{Results comparison on face paintings by Raphael in 10-shot setting. We test the results on FFHQ, which is the source domain of the other three models. GPD presents a competitive style compared to other methods and the most consistent content with the source image (Input). }
    \label{fig:ap_r}
    \end{figure*}

\begin{figure*}[t]
    \centering
    \includegraphics[width=1.0\linewidth]{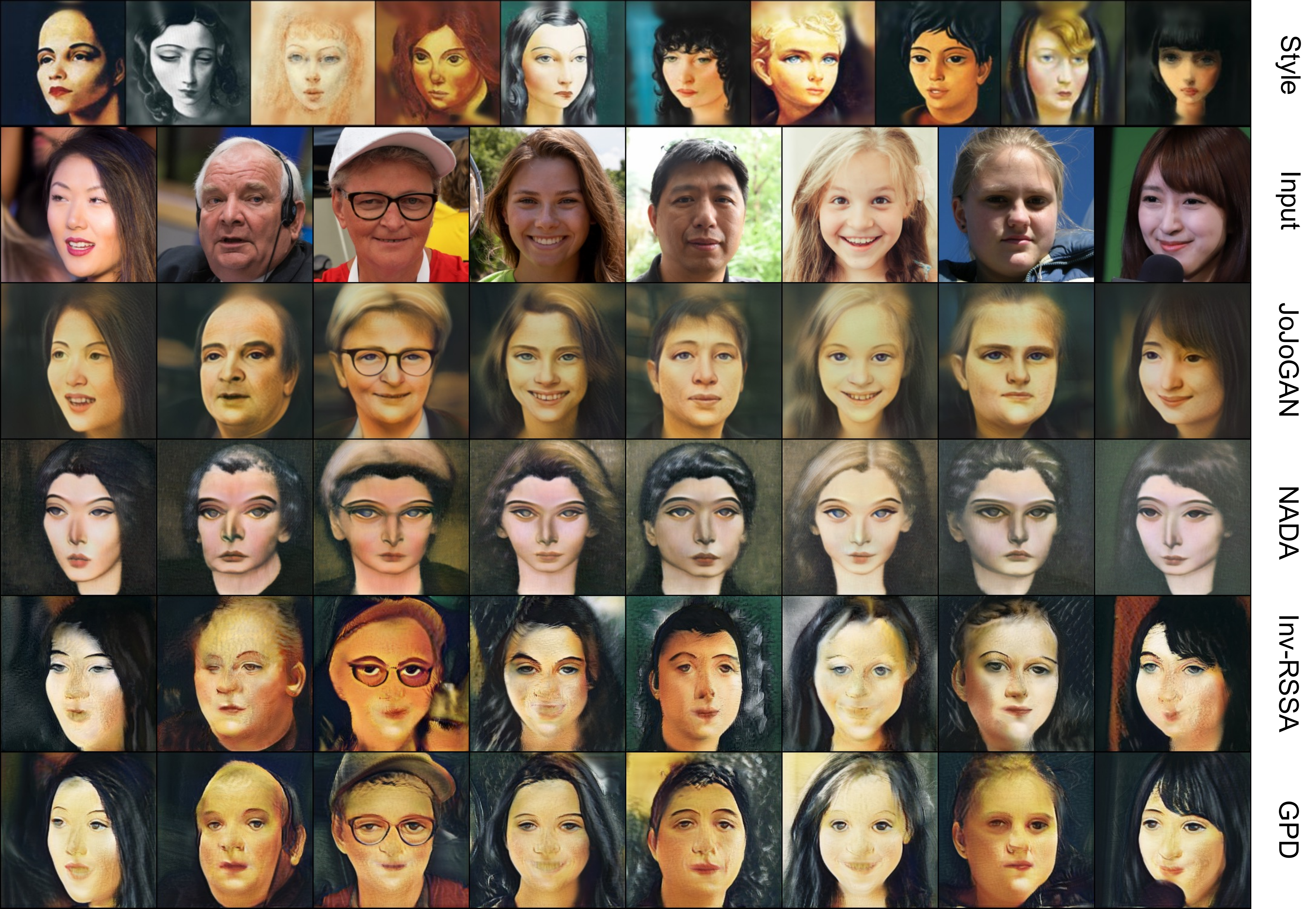}
    \caption{Results comparison on face paintings by Moise Kisling in 10-shot setting. We test the results on FFHQ, which is the source domain of the other three models. GPD presents a competitive style compared to other methods and the most consistent content with the source image (Input).}
    \label{fig:ap_c}
    \end{figure*}

\begin{figure*}[t]
    \centering
    \includegraphics[width=1.0\linewidth]{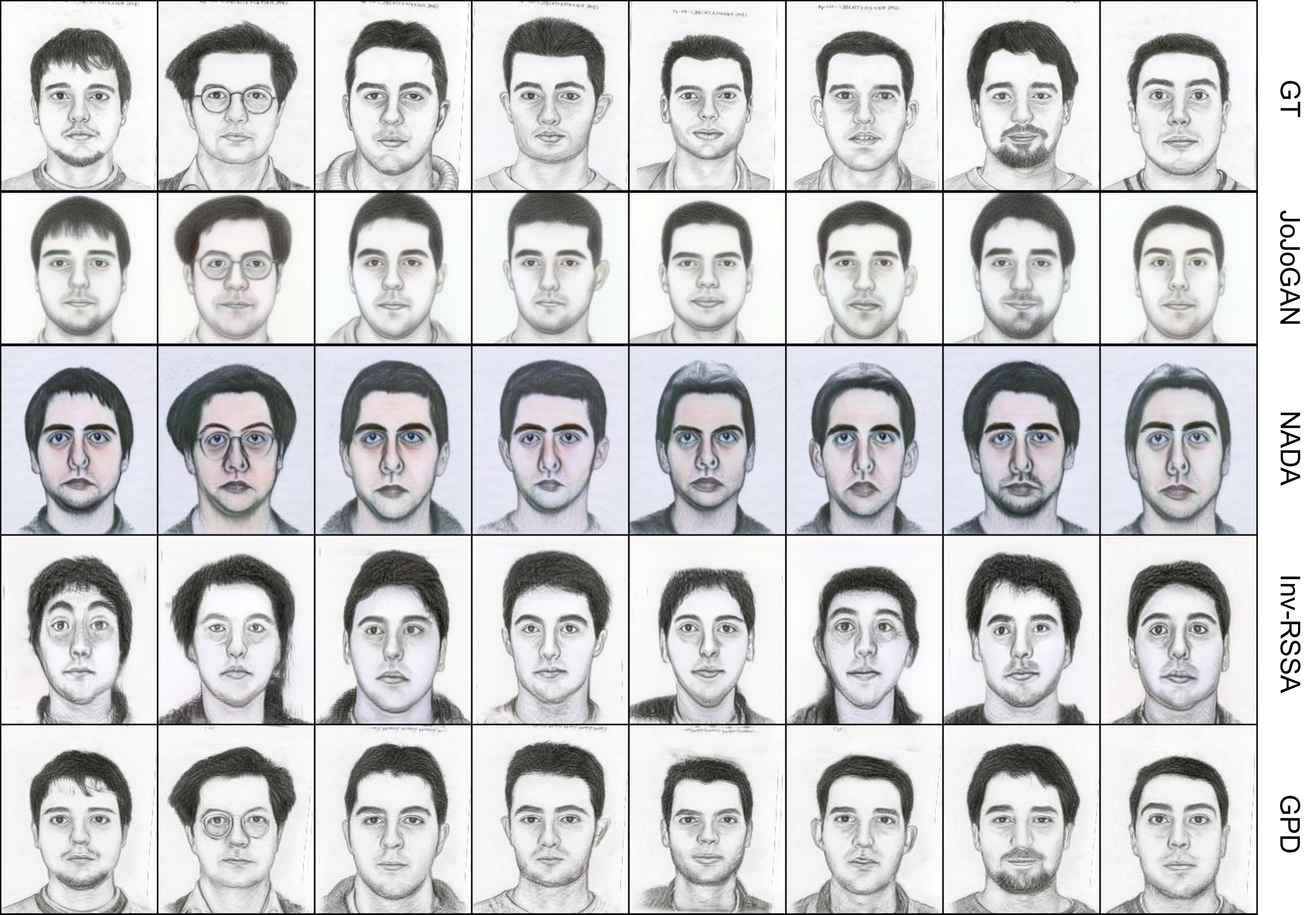}
    \caption{Result comparison on AR in 10-shot setting. GPD presents the best style and the most consistent content with ground truth (GT). This shows that GPD performance is better than other three models in the domain of training data itself.}
    \label{fig:ap_ar}
    \end{figure*}

\begin{figure*}[t]
    \centering
    \includegraphics[width=1.0\linewidth]{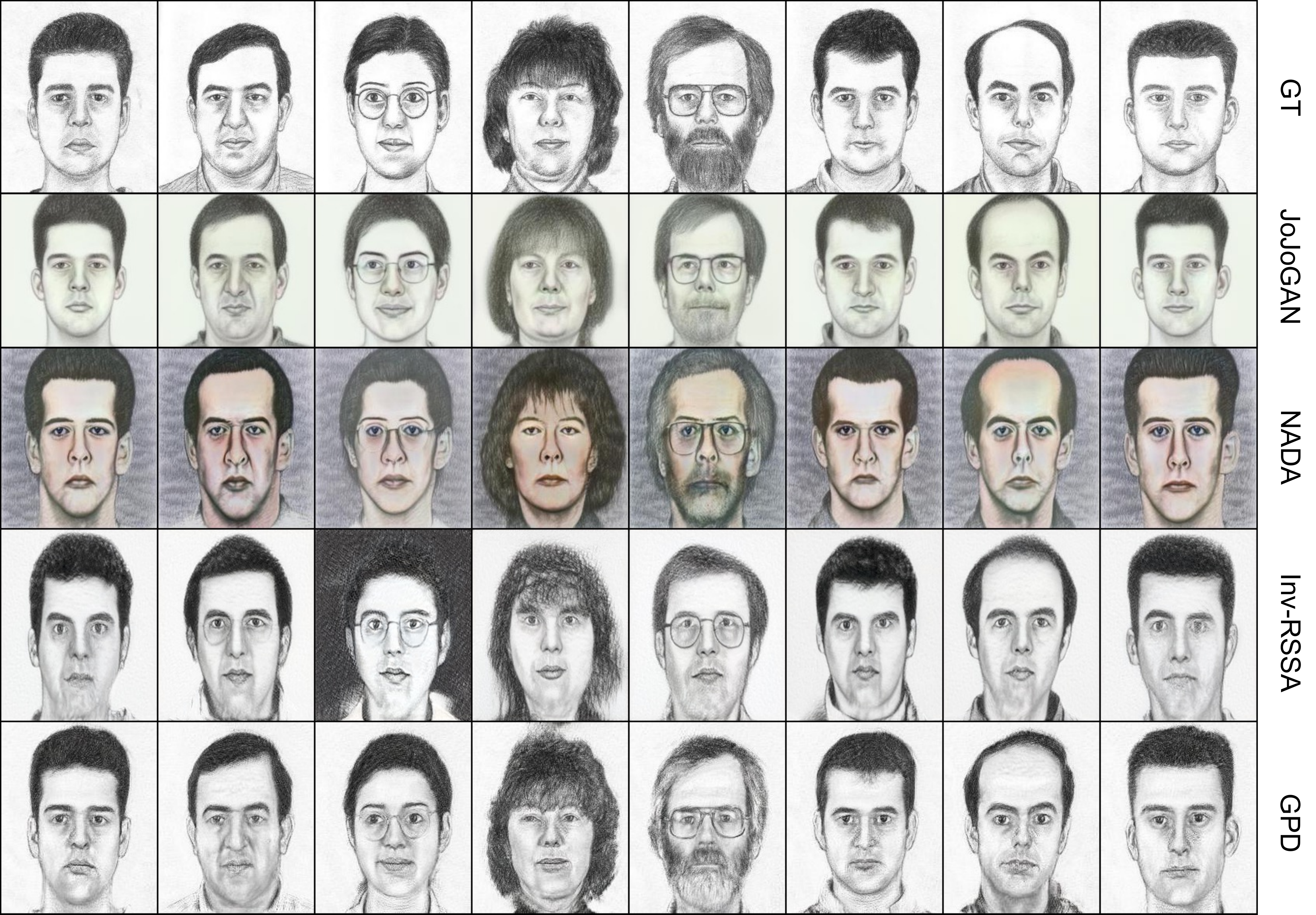}
    \caption{Result comparison on XM2VTS in 10-shot setting. GPD presents the best style and the most consistent content with ground truth (GT).}
    \label{fig:ap_xm2vts}
    \end{figure*}

Figure \ref{fig:ap_a}-\ref{fig:ap_xm2vts} show the result comparison between GPD and the other three methods in the 10-shot face image translation task on different datasets.
For Figure \ref{fig:ap_a}-\ref{fig:ap_r}, we test the results of face paintings by Amedeo Modigliani, Moise Kisling, and Raphael on FFHQ.
FFHQ is the source domain of the other three models. 
GPD presents a competitive style compared to other methods and the most consistent content with the source image (Input), which also shows the generalization performance of GPD.
For Figure \ref{fig:ap_ar}-\ref{fig:ap_xm2vts}, GPD presents the best style and consistent content with ground truth (GT). It shows that GPD performance is better than the other three models in the domain of training data itself.

\begin{figure*}[t]
    \centering
    \includegraphics[width=1.0\linewidth]{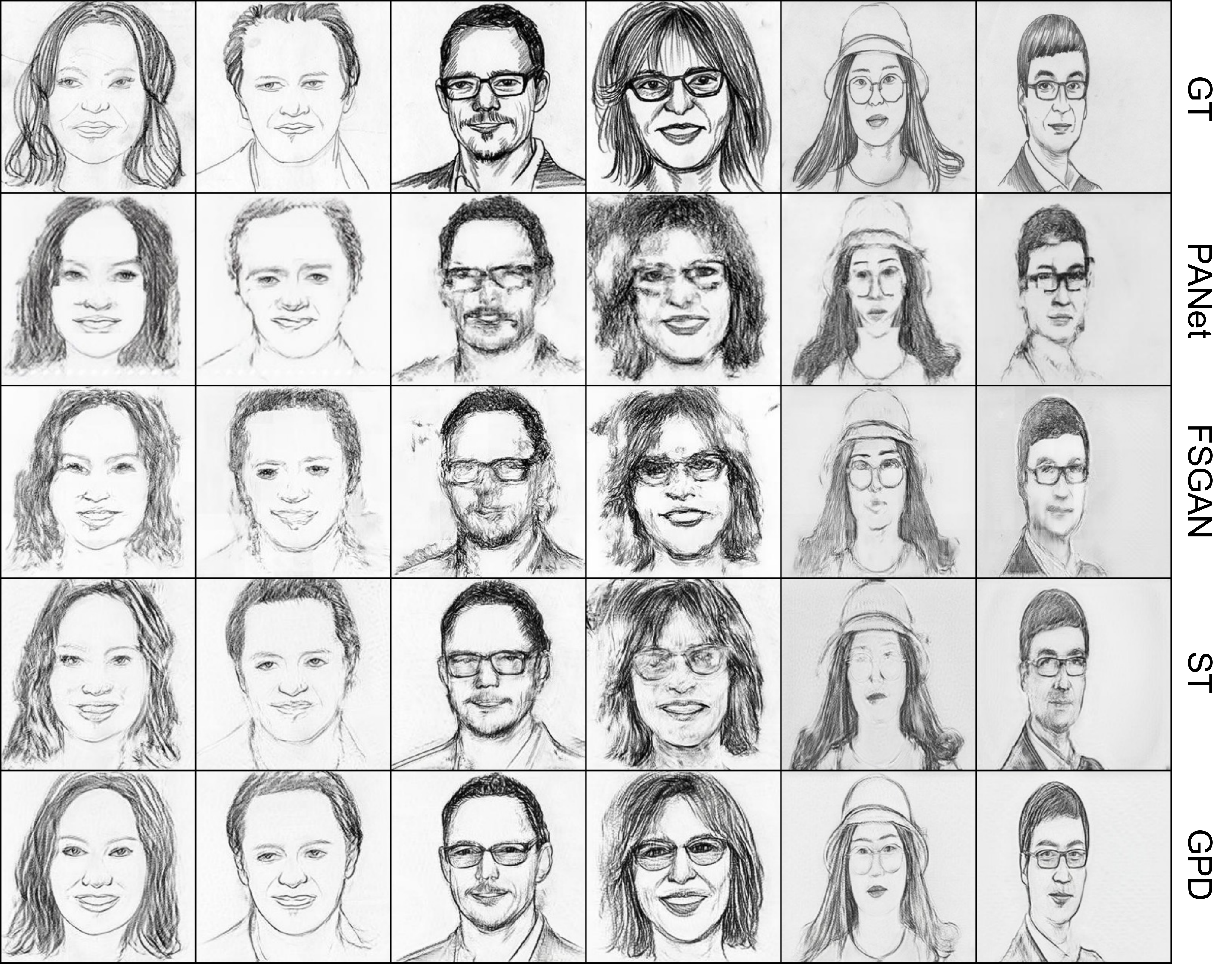}
    \caption{Results comparison with three sketch synthesis networks
    in 20-shot setting. We display two samples for each style.}
    \label{fig:ap_compare_fss}
    \end{figure*}

Figure \ref{fig:ap_compare_fss} shows the result comparison between GPD and the other three facial-sketch synthesis networks in 20-shot setting. GPD preserves a complete structure in few-shot settings, which is beyond the reach of other sketch synthesis networks. 

\begin{table}[b]
    \centering
    \begin{tabular}{l l c c c c c}   
        \toprule
        \multirow{2}*{Style}  & \multirow{2}*{Data Scale} &\multicolumn{2}{c}{LPIPS$\downarrow$} &\multicolumn{2}{c}{SSIM$\uparrow$}      \\\cmidrule(r){3-4} \cmidrule(r){5-6}
            & &BL &Aug &BL &Aug \\
        \midrule
        \multirow{3}*{S1} & 6.25$\%$   & 0.322 & \textbf{0.248} & 0.471 & \textbf{0.503}\\
                          & 12.5$\%$   & 0.284 & \textbf{0.246} & 0.502 & \textbf{0.521}  \\
                          & 25$\%$     & 0.257 & \textbf{0.242} & 0.505 & \textbf{0.519}  \\
                          & 50$\%$     & 0.242 & \textbf{0.233} & 0.507 & \textbf{0.519}  \\
                          & 100$\%$    & \textbf{0.231} & 0.236 & 0.517 & \textbf{0.521}  \\
            
        \midrule
        \multirow{3}*{S2} & 6.25$\%$   & 0.333 & \textbf{0.285} & 0.319 & \textbf{0.348}  \\
                          & 12.5$\%$   & 0.289 & \textbf{0.254} & 0.337 & \textbf{0.345}  \\
                          & 25$\%$     & 0.263 & \textbf{0.259} & 0.348 & \textbf{0.355}  \\
                          & 50$\%$     & 0.243 & \textbf{0.237} & 0.353 & \textbf{0.357}  \\
                          & 100$\%$    & \textbf{0.239} & 0.243 & 0.364 & \textbf{0.366}  \\                        
        \midrule
        \multirow{3}*{S3} & 6.25$\%$   & 0.326 & \textbf{0.249} & 0.463 & \textbf{0.545}  \\
                          & 12.5$\%$   & 0.296 & \textbf{0.258} & 0.486 & \textbf{0.536}  \\
                          & 25$\%$     & 0.281 & \textbf{0.243} & 0.483 & \textbf{0.541}  \\
                          & 50$\%$     & 0.258 & \textbf{0.233} & 0.509 & \textbf{0.554}  \\
                          & 100$\%$    & 0.252 & \textbf{0.232} & 0.497 & \textbf{0.542}  \\     
        \bottomrule
        \end{tabular}
    \caption{Comparison of results with and without using our augmentation method on U-net under different data scales. S1, S2, S3: the three styles of FS2K; BL: baseline; Aug: using our method for augmentation.}
    \label{tab:ap_plot1}
\end{table}

\begin{table}[b]
    \centering
    \begin{tabular}{l l c c c c c}   
        \toprule
        \multirow{2}*{Style}  & \multirow{2}*{Data Scale} &\multicolumn{2}{c}{LPIPS$\downarrow$} &\multicolumn{2}{c}{SSIM$\uparrow$}      \\\cmidrule(r){3-4} \cmidrule(r){5-6}
            & &BL &Aug &BL &Aug \\
        \midrule
        \multirow{3}*{S1} & 6.25$\%$   & 0.389 & \textbf{0.291} & 0.374 & \textbf{0.480}\\
                          & 12.5$\%$   & 0.367 & \textbf{0.273} & 0.404 & \textbf{0.507}  \\
                          & 25$\%$     & 0.334 & \textbf{0.279} & 0.446 & \textbf{0.513}  \\
                          & 50$\%$     & 0.319 & \textbf{0.278} & 0.439 & \textbf{0.508}  \\
                          & 100$\%$    & 0.298 & \textbf{0.278} & 0.492 & \textbf{0.507}  \\
            
        \midrule
        \multirow{3}*{S2} & 6.25$\%$   & 0.403 & \textbf{0.314} & 0.223 & \textbf{0.317}  \\
                          & 12.5$\%$   & 0.353 & \textbf{0.292} & 0.269 & \textbf{0.344}  \\
                          & 25$\%$     & 0.319 & \textbf{0.294} & 0.297 & \textbf{0.336}  \\
                          & 50$\%$     & 0.305 & \textbf{0.288} & 0.320 & \textbf{0.348}  \\
                          & 100$\%$    & 0.307 & \textbf{0.290} & 0.294 & \textbf{0.350}  \\                        
        \midrule
        \multirow{3}*{S3} & 6.25$\%$   & 0.319 & \textbf{0.274} & 0.497 & \textbf{0.552}  \\
                          & 12.5$\%$   & 0.289 & \textbf{0.269} & 0.531 & \textbf{0.570}  \\
                          & 25$\%$     & 0.267 & \textbf{0.263} & 0.540 & \textbf{0.573}  \\
                          & 50$\%$     & \textbf{0.256} & 0.263 & 0.560 & \textbf{0.564}  \\
                          & 100$\%$    & \textbf{0.253} & 0.258 & 0.569 & \textbf{0.585}  \\     
        \bottomrule
        \end{tabular}
    \caption{Comparison of results with and without using our augmentation method on Pix2Pix under different data scales. S1, S2, S3: the three styles of FS2K; BL: baseline; Aug: using our method for augmentation.}
    \label{tab:ap_plot2}
\end{table}



\end{appendices}
\end{document}